\documentclass{article}
\usepackage{ijcai25}
\usepackage{amssymb}
\usepackage{times}
\usepackage{soul}
\usepackage{url}
\usepackage[hidelinks]{hyperref}
\usepackage[utf8]{inputenc}
\usepackage[small]{caption}
\usepackage{graphicx}
\usepackage{amsmath}
\usepackage{amsthm}
\usepackage{booktabs}
\usepackage{algorithm}
\usepackage{algorithmic}
\usepackage[switch]{lineno}

\usepackage{multirow}
\usepackage{subfig}

\title{Top-Down Guidance for Learning Object-Centric Representations}

\author{
	Junhong Zou$^{1,2}$
	\and
	Xiangyu Zhu$^{1,2}$\thanks{Corresponding author} \and
	Zhaoxiang Zhang$^{1,2,3}$\And
	Zhen Lei$^{1,2,3,4}$ \\
	\affiliations
	$^1$MAIS, Institute of Automation, Chinese Academy of Sciences, Beijing, China \\
	$^2$School of Artificial Intelligence, University of Chinese Academy of Sciences, Beijing, China\\
	$^3$CAIR, HKSIS, Chinese Academy of Sciences, Hong Kong, China\\
	$^4$School of Computer Science and Engineering, the Faculty of Innovation Engineering, M.U.S.T, Macau, China\\
	\emails
	\{zoujunhong2022, xiangyu.zhu, zhaoxiang.zhang, zhen.lei\}@ia.ac.cn
}

\begin{document}
	
	\maketitle
	
	\begin{figure*}[ht]
		\centering
		\includegraphics[width=1\linewidth]{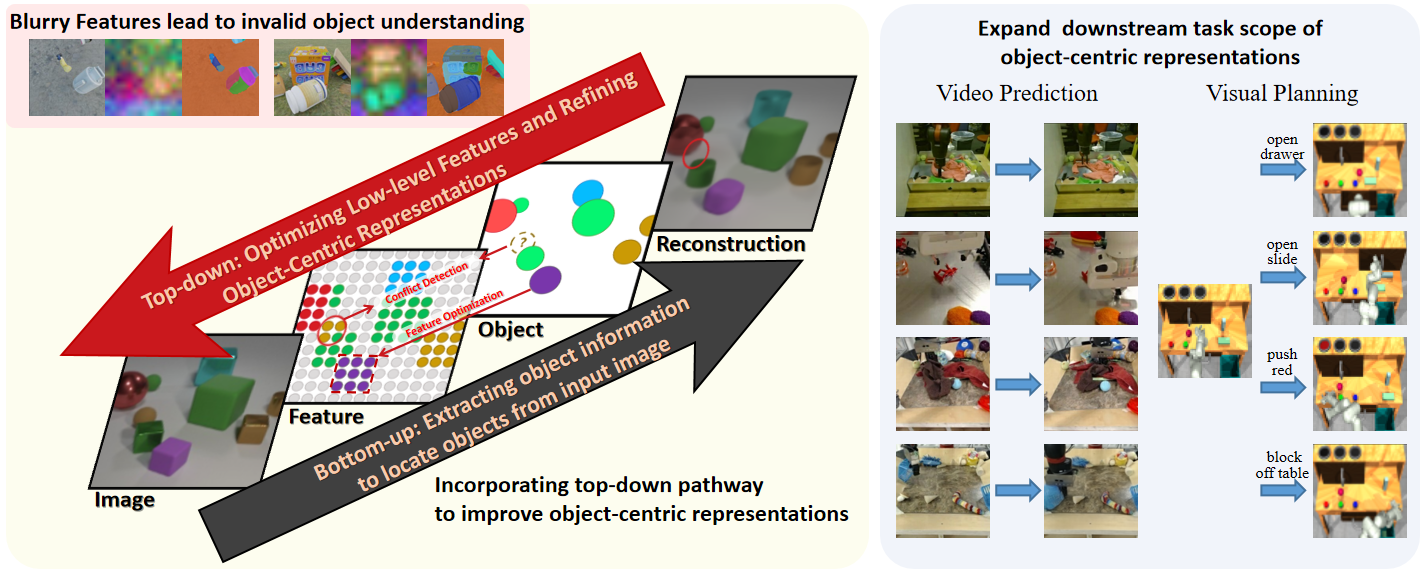}
		
		\caption{\textbf{Overview}. Observing that existing object-centric models fail to learn distinguishable features, limiting the models' object understanding ability, we propose TDGNet that introduces a top-down pathway to help optimize the low-level features output by the backbone and refine object-centric representation. Furthermore, we apply the refined representations to the field of robotics to demonstrate that TDGNet can adapt to a broad range of tasks and scenarios.}
		\label{fig: summary}
		\vspace{-0.8em}
	\end{figure*}
	
	\begin{abstract}
			Humans' innate ability to decompose scenes into objects allows for efficient understanding, predicting, and planning. In light of this, Object-Centric Learning (OCL) attempts to endow networks with similar capabilities, learning to represent scenes with the composition of objects. However, existing OCL models only learn through reconstructing the input images, which does not assist the model in distinguishing objects, resulting in suboptimal object-centric representations. This flaw limits current object-centric models to relatively simple downstream tasks. To address this issue, we draw on humans’ top-down vision pathway and propose Top-Down Guided Network (TDGNet), which includes a top-down pathway to improve object-centric representations. During training, the top-down pathway constructs guidance with high-level object-centric representations to optimize low-level grid features output by the backbone. While during inference, it refines object-centric representations by detecting and solving conflicts between low- and high-level features. We show that TDGNet outperforms current object-centric models on multiple datasets of varying complexity. In addition, we expand the downstream task scope of object-centric representations by applying TDGNet to the field of robotics, validating its effectiveness in downstream tasks including video prediction and visual planning. Code will be available at \href{https://github.com/zoujunhong/RHGNet}{https://github.com/zoujunhong/RHGNet}.
		
	\end{abstract}

	\section{Introduction}
	\label{introduction}
	
	Humans are skilled at decomposing visual scenes into the compositions of objects~\cite{kahneman1992reviewing}, which is crucial for humans' efficient understanding, predicting, and planning. Inspired by this property, Object-Centric Learning (OCL) seeks to achieve human-like representations in neural networks. Specifically, models are trained in a self-supervised manner to represent visual signals, such as images or videos with a set of latent vectors which are referred to as `slots'~\cite{greff2019multi,https://doi.org/10.48550/arxiv.2006.15055}. Ideally, each slot corresponds to an object in the scene.
	
	Previous methods \cite{https://doi.org/10.48550/arxiv.2006.15055,jia2023improving,slate,jiang2023objectcentric} have achieved considerable progress in OCL. However, we observe that the performance of existing methods is highly unstable across different scenarios. For example, in Figure \ref{fig: summary}, we present the cases where current models show sup-optimal object understanding and decompose scenes in an inferior manner: some objects are missed or split into parts. This expresses a concern that current models highly rely on the inductive biases of model structures and visual scenes to learn object-centric representations, posing challenges for adapting to various scenarios and tackling downstream tasks.
	
	We attribute this issue to the fact that current OCL models typically adopt an auto-encoding paradigm that learns object-centric representations by encoding images into slots and using these slots to reconstruct images. However, reconstruction loss does not tell apart objects: models do not need to decompose the scene according to objects. As a result, these models fail to learn distinguishable features at the backbone. For instance, in Figure \ref{fig: summary} and \ref{fig: variance}, we visualize the output features of the backbone when the object-centric models are trained solely with reconstruction loss. It is observed that these features are blurry and fail to align with the edges of objects. Moreover, the features of small objects are particularly challenging to distinguish from the background. Such indistinguishable feature makes it difficult for the model to decide how to assign features to slots, resulting in suboptimal object-centric representations.
	
	
	To address this issue, we propose to guide the model by an additional top-down pathway. This coincides with humans' perceptual learning process, where theories about human vision~\cite{hochstein2002view,wolfe2021guided2} argue that humans first learn concepts at high-level consciousness and then use the high-level perception to guide the learning of low-level neurons. Drawing on this mechanism, we propose \textbf{Top-Down Guided Network} (\textbf{TDGNet}) that introduces the top-down pathway to use the high-level representations (i.e., the slots) to guide the low-level features output by the backbone. Specifically, we obtain a high-level guidance by weighted summing the slots according to their masks, and then introduce a projection network to predict this guidance signal using low-level features. In this way, the model tends to cluster low-level features belonging to the same slots, and otherwise keep them apart, thus making low-level features more distinguishable. Moreover, we extend this concept to the inference phase, introducing a conflict detection method designed to refine object-centric representations during inference: when a feature's prediction is far from all existing slots or close to multiple slots, it may represent a suboptimal perception such as missing objects or splitting objects into parts, which we call a conflict. We detect and solve such conflicts by adding or merging slots, thus refining the object-centric representations.

	We evaluate TDGNet and compare it with current SOTA models on multiple tasks. We first introduce CLEVRTex~\cite{https://doi.org/10.48550/arxiv.2111.10265}, MOVi-C~\cite{movi} and COCO to evaluate the object-centric representations, where TDGNet outperforms current SOTA models in terms of common object discovery metrics. Furthermore, we expand the downstream task scope of TDGNet by applying it to the field of robotics. We introduce RoboNet~\cite{robonet} and $\mathrm{VP^2}$~\cite{vp2}, to evaluate TDGNet with downstream tasks including video prediction and visual planning, demonstrating that TDGNet adapts well to these tasks.
	
	To sum up, our contributions are summarized as follows:
	\begin{itemize}
		
		\item[-] Drawing on the top-down visual pathway of humans, we propose TDGNet, which incorporates a top-down pathway that constructs high-level guidance to optimize the low-level features, thereby improving object-centric representations.
		
		\item[-] Based on the top-down pathway, we propose conflict detection to discover perceptual errors and further refine object-centric representations during inference. 
		
		\item[-] We demonstrate the SOTA performance of TDGNet in object-centric representation tasks. Besides, we introduce it into more complex robotic scenarios and verify that it works effectively in the video prediction and visual planning task.
	\end{itemize}
	
	\vspace{-0.5em}
	\section{Related Work}
	\noindent  \textbf{Object-Centric Learning.} Most current OCL methods follow an auto-encoding paradigm that first encodes input signals into several slots and reconstructs the original signal with these slots. Earlier works, including IODINE~\cite{greff2019multi}, MONet~\cite{burgess2019monet} and GENESIS~\cite{genesis}, accomplish this task by using multiple encoder-decoder structures. Slot-Attention~\cite{https://doi.org/10.48550/arxiv.2006.15055} proposed an iterative attention method that allows slots to compete for input image segments and conduct segmentation. A critical issue of current OCL methods is how to generalize to more complex scenes. BO-QSA~\cite{jia2023improving}, I-SA~\cite{https://doi.org/10.48550/arxiv.2207.00787} and InvariantSA~\cite{biza2023invariant} focus on query optimization, which uses learnable parameters to initialize slots. SLATE~\cite{slate} and LSD~\cite{jiang2023objectcentric} attempt to improve the decoder structure, introducing transformer-based and diffusion-based decoders to enhance the model's reconstruction ability. DINOSAUR~\cite{seitzer2022bridging} proposes that the simple reconstruction task is insufficient to distinguish objects and replaces the reconstruction objective with the output feature of DINO~\cite{caron2021emerging}.
	
	\begin{figure*}[t]
		\centering
		\subfloat[Top-down Guidance (TDG) during training.]{\includegraphics[width=1.8\columnwidth]{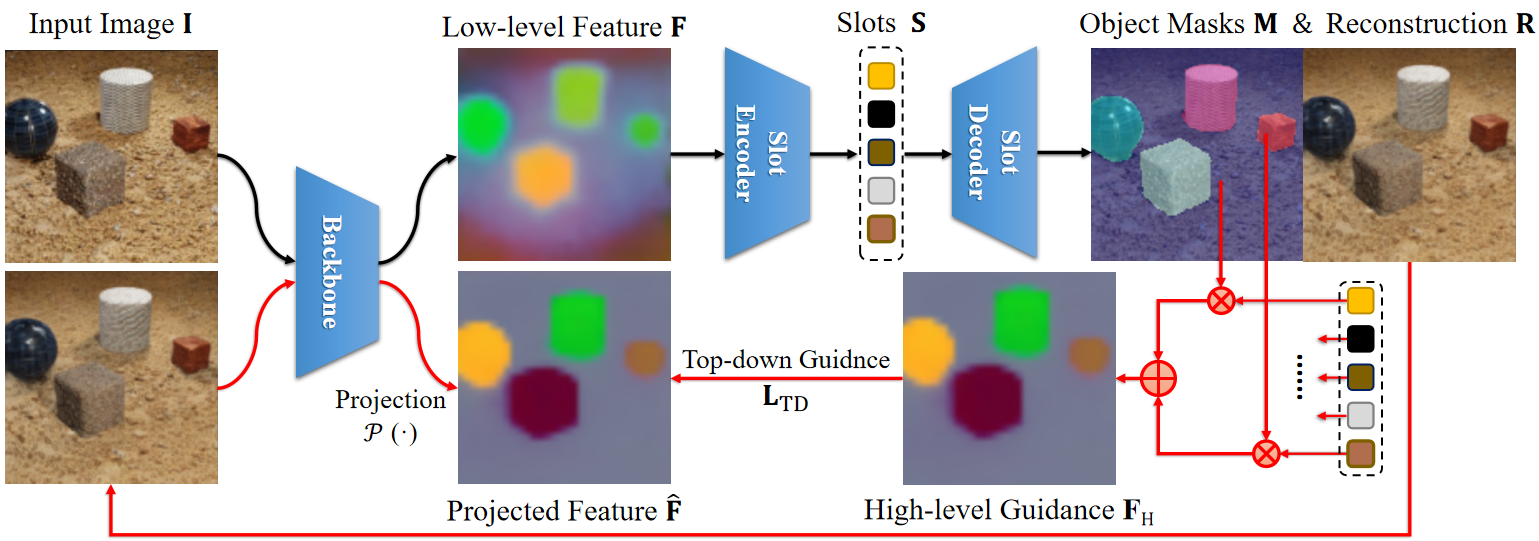}}\hspace{3pt}
		\\
		\subfloat[Conflict detection (CD) for refining representations during inference.]{\includegraphics[width=1.8\columnwidth]{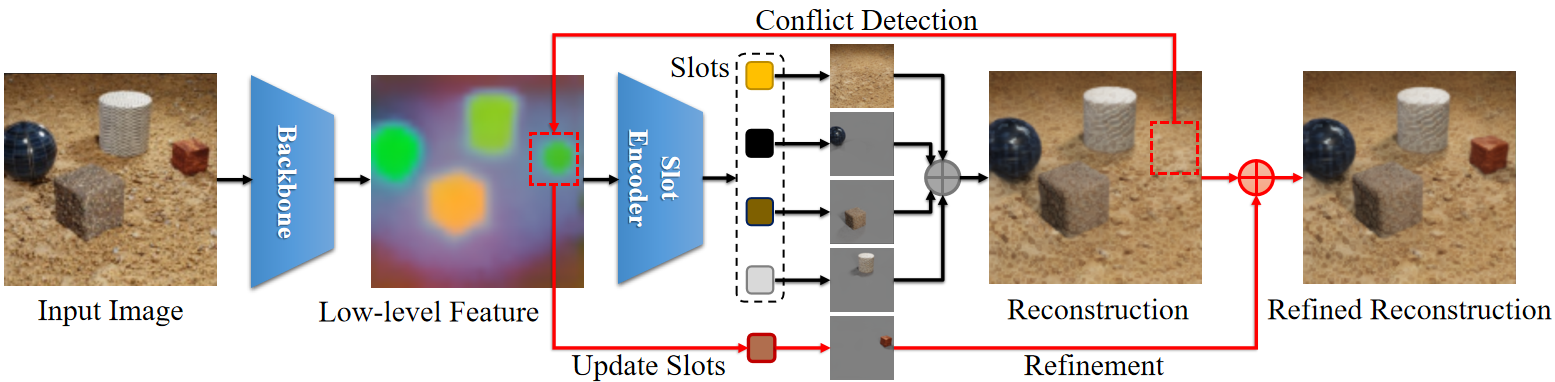}}
		\caption{\textbf{Architecture of Top-Down Guided Network (TDGNet)}. The proposed TDGNet acquires the initial perception through the bottom-up pathway (black arrow) and refines its object-centric representations with the top-down pathway (red arrow). \textbf{(a)} During training, TDGNet uses slots and object masks to guide the backbone to produce distinguishable low-level features. \textbf{(b)} During Inference, TDGNet refines the slots by detecting conflicts between its slots and the low-level features.}
		\vspace{-0.5em}
		\label{fig: architecture}
	\end{figure*}
	
	\noindent  \textbf{Top-down connections in human vision.} Human brain transmits high-level semantic information to low-level neurons~\cite{beck2009top}, resulting in a biased competition between different objects to control attention. In addition, the brain can receive task-relevant information and inhibit irrelevant neurons to improve the efficiency of completing tasks~\cite{karimi2017invariant}. Reverse hierarchy theory~\cite{hochstein2002view,ahissar2004reverse} generalizes neural connections in the human brain into two functional pathways: the bottom-up pathway works implicitly, acquiring the gist of the scene rapidly, and the top-down pathway returns to lower-level neurons to bring detailed information into consciousness.
	
	\noindent  \textbf{Top-down connections in Neural Networks.} It has long been explored to incorporate top-down feedback connections into neural networks.~\cite{liang2015recurrent} introduce recurrent connections into convolutional networks.~\cite{wen2018deep} propose to achieve predictive coding through a network with feedback connections. Recent works have utilized a top-down pathway to solve different visual or multi-modal tasks, including semantic segmentation~\cite{yin2022transfgu,liu2024primitivenet}, visual saliency~\cite{ramanishka2017top} and vision question \& answering~\cite{anderson2018bottom}. Most of these models integrate features from multiple layers through a parameterized network module, introducing additional depth into the network through feedback connections. However, there is no evidence that they achieve the visual functions of human feedback connections.


	
	\section{Method}
	\subsection{Preliminary: Auto-encoding-based Object-Centric Models}
	\label{BU}
	The architecture of TDGNet is shown in Figure \ref{fig: architecture}. A typical auto-encoding-based object-centric model serves as our \textbf{bottom-up pathway}: the backbone $\mathcal{E}_B$ first extract the low-level features $\mathbf{F} \in \mathbb{R}^{C_f\times H/s \times W/s}$ from the input image $\mathbf{I} \in \mathbb{R}^{3\times H \times W}$. Then a slot encoder $\mathcal{E}_S$ encodes $\mathbf{F}$ into $K$ slots $\mathbf{S} \in \mathbb{R}^{K\times C_s}$, which is regarded as the high-level representations. Finally, a slot decoder $\mathcal{D}_S$ decodes slots into reconstructions $\mathbf{R} \in \mathbb{R}^{3\times H \times W}$ and object masks $\mathbf{M} \in \mathbb{R}^{K \times H \times W}$. Formally, the bottom-up pathway is described as:
	
	\begin{equation}
		\left\{
		\begin{aligned}
			&\mathbf{F} = \mathcal{E}_B(\mathbf{I}),\\
			&\mathbf{S} = \mathcal{E}_S(\mathbf{F}),\\
			&\mathbf{R}, \mathbf{M} = \mathcal{D}_S(\mathbf{S}).\\
		\end{aligned}
		\right.
	\end{equation}
	The bottom-up pathway is optimized by reconstructing input images. Here we use a combination of L1 loss and perceptual loss (LPIPS)~\cite{zhang2018unreasonable} for optimization. The reconstruction loss is written as: 
	
	\begin{equation}
		\mathbf{L}_\mathrm{rec} := \Vert \mathbf{R}-\mathbf{I}\Vert_1 + \mathrm{LPIPS}(\mathbf{R}, \mathbf{I})
	\end{equation}
	
	\subsection{Learning Object-Centric Representations with Top-down Pathway}
	
	We introduce a top-down pathway to refine the object-centric representations, which works in two ways: during training, it constructs guidance with high-level slots to optimize low-level features output by the backbone; during inference, it refines the slots by detecting and solving conflicts between slots and low-level features.
	
	\begin{table*}[!ht]
		\small
		\begin{center}

			\setlength{\tabcolsep}{0.3mm}
			\begin{tabular}{l c c c c c c c c c c c c c} 
				\toprule[2pt]
				\makebox[0.25\textwidth][c]{\multirow{2}{*}{\textbf{Model}}}  & 
				\multicolumn{3}{c}{\makebox[0.05\textwidth][c]{CLEVRTex}} & 
				\multicolumn{3}{c}{\makebox[0.05\textwidth][c]{CAMO}} & 
				\multicolumn{3}{c}{\makebox[0.05\textwidth][c]{OOD}} & 
				\multicolumn{4}{c}{\makebox[0.05\textwidth][c]{MOVi-C}} \\
				\cmidrule(r){2-4}\cmidrule(r){5-7}\cmidrule(r){8-10}\cmidrule(r){11-14}
				& $\uparrow$ARI-FG & $\uparrow$mIoU & $\downarrow$MSE & $\uparrow$ARI-FG & $\uparrow$mIoU & $\downarrow$MSE& $\uparrow$ARI-FG & $\uparrow$mIoU & $\downarrow$MSE& $\uparrow$ARI-FG & $\uparrow$mIoU & $\uparrow$mBO & $\downarrow$MSE \\
				\hline
				SLATE~\cite{slate} &45.4&49.5&498&43.5&37.7&349&46.5&35.4& 550 &  49.5 & 37.8 & 39.4 & 526 \\
				LSD~\cite{jiang2023objectcentric} & 64.4 & 62.5 & 237 & 62.6 & 60.8 & 245 & 58.9 & 56.4 & 492 & 52.3 & 44.1 & 45.6 & 661 \\
				BO-QSA~\cite{jia2023improving} &  80.5 & 46.7 & 268 & 72.6 & 41.5 & 246 & 72.5 & 37.1 & 805 & 52.9 & 33.1 & 36.4 & 157 \\
				InvariantSA~\cite{biza2023invariant} &  92.9 & 72.4 & 177 & 86.2 & 65.6 & 196 & \underline{84.4} & 66.7 & 578 & 35.7 & 26.0 & 26.9 & 484 \\
				DINOSAUR~\cite{seitzer2022bridging} & 88.9 & 52.6 & - & 83.5 & 51.3 & - & 83.1 & 51.9 & - & \underline{67.8} & 31.2 & 38.2 & - \\
				\hline
				\textbf{TDGNet} & \underline{94.2} & \underline{80.3} & \underline{65} & \underline{88.9} & \underline{76.3}  & \underline{82} & 84.1 & \underline{69.6} & \underline{302} & 61.2 & \underline{52.9} & \underline{53.5} & \underline{151} \\
				\textbf{+CD} & \textbf{94.8} & \textbf{80.5} & \textbf{63} & \textbf{89.5} & \textbf{77.0} & \textbf{74} &  \textbf{84.8}  & \textbf{71.9} & \textbf{291} & \textbf{68.5} & \textbf{55.6}  & \textbf{57.1} & \textbf{148} \\
				\bottomrule[2pt]
			\end{tabular}
			\caption{Model performance comparison on CLEVRTex and MOVi-C. CAMO and OOD represent CLEVRTex-CAMO and CLEVRTex-OOD where models trained on CLEVRTex are directly evaluated without finetune.}
			
			\label{tab1}
		\end{center}
		
	\end{table*}
	
	\begin{table}[!ht]
		\small
		\begin{center}

			\setlength{\tabcolsep}{0.7mm}
			\begin{tabular}{l c c c} 
				\toprule[2pt]
				\makebox[0.26\textwidth][c]{\textbf{COCO}} &
				\makebox[0.05\textwidth][c]{ARI-FG} & 
				\makebox[0.05\textwidth][c]{$\mathrm{mBO^i}$} & 
				\makebox[0.05\textwidth][c]{mIoU} \\
				\hline
				\multicolumn{4}{c}{\makebox[0.05\textwidth][c]{MLP-based methods}} \\
				\hline
				SA~\cite{https://doi.org/10.48550/arxiv.2006.15055} 
				& 17.5 & 18.2 & 12.2 \\
				BO-QSA~\cite{jia2023improving} 
				& 35.7 & 26.0 & 26.9 \\
				DINOSAUR-mlp~\cite{seitzer2022bridging} 
				& 40.5 & 27.7 & 26.4 \\
				DINOSAUR-mlp + DINOv2 
				& \underline{42.9} & \underline{28.9} & \underline{27.3} \\
				\textbf{TDGNet (ours)} 
				& \textbf{45.0} & \textbf{29.6} & \textbf{28.5} \\
				\hline
				\multicolumn{4}{c}{\makebox[0.05\textwidth][c]{Transformer/Diffusion-based methods}} \\
				\hline
				SLATE~\cite{slate} 
				& 23.2 & 20.2 & 19.3 \\
				LSD~\cite{jiang2023objectcentric} 
				& \underline{37.0} & 34.8 & 32.2 \\
				DINOSAUR-tf~\cite{seitzer2022bridging} 
				& 32.3 & 32.0 & 30.0 \\
				SPOT~\cite{Kakogeorgiou_2024_CVPR} 
				& \underline{37.0} & \underline{35.0} & \underline{33.0} \\
				\textbf{TDGNet (ours)} 
				& \textbf{37.3} & \textbf{35.6} & \textbf{33.2} \\
				\bottomrule[2pt]
			\end{tabular}
			\caption{Unsupervised object discovery result on COCO. Higher is better for all the metrics.}
			
			\label{tab2}
		\end{center}
	\end{table}

	\subsubsection{Top-down Guidance during Training}
	
	We first introduce how the top-down pathway constructs the \textbf{Top-Down Guidance} (\textbf{TDG}) to optimize low-level features during training. As shown in Figure \ref{fig: architecture}(a), the bottom-up pathway provides initial object-centric representations with slots. The top-down pathway utilizes these slots to construct guidance for training. Ideally, it makes features from the same slot more similar than those from different ones. 
	
	The guidance is constructed through high-level representations including object masks $\mathbf{M}$ and slots $\mathbf{S}$. Formally, the high-level guidance $\mathbf{F}_\mathrm{H}$ is obtained through the sum of $\mathbf{S}$, weighted by $\mathbf{M}$ at each spatial location:
	\begin{equation}
		\mathbf{F}_\mathrm{H} = \mathcal{SG}(\mathrm{Sum}(\mathbf{S} * \mathbf{M}, \mathrm{axis=slots})).
	\end{equation}
	Here $\mathcal{SG}$ represents the stop-gradient operation. We stop the gradient of high-level signals (namely $\mathbf{M}$ and $\mathbf{S}$) so that the guidance only works on the low-level features.
	
	Subsequently, we introduce a projection network $\mathcal{P}$ that uses low-level features to predict the high-level guidance $\mathbf{F}_\mathrm{H}$. In this way, the low-level features are required to predict their corresponding slots. Considering that $\mathbf{F}_\mathrm{H}$ are produced with the slots $\mathbf{S}$, it provides more accurate object regions for the reconstructed images $\mathbf{R}$ than the input images $\mathbf{I}$. Therefore, we let $\mathcal{P}$ project the low-level features extracted from $\mathbf{R}$ for more accurate guidance. Formally, we re-input $\mathbf{R}$ into the backbone $\mathcal{E}_B$ to extract its low-level features $\widehat{\mathbf{F}}$:
	\begin{equation}
		\widehat{\mathbf{F}} = \mathcal{E}_B(\mathcal{SG}(\mathbf{R})).
	\end{equation}
	Finally, we use the projection network $\mathcal{P}$ to predict $\mathbf{F}_\mathrm{H}$ with $\widehat{\mathbf{F}}$. In our method, slots $\mathbf{S}$ are normalized to unit length, and the distance is measured through cosine similarity $\mathrm{CosSim}$. The top-down guidance loss $\mathbf{L}_\mathrm{TD}$ is formulated as below:
	\begin{equation}
		\label{eqa:1}
		\mathbf{L}_\mathrm{TD} := 1 - \mathrm{CosSim}(\mathcal{P}(\widehat{\mathbf{F}}), \mathbf{F}_\mathrm{H}).
	\end{equation}
	Overall, TDGNet is trained by $\mathbf{L}$, the weighted sum of the reconstruction loss and the top-down guidance loss:
	\begin{equation}
		\mathbf{L} = \mathbf{L}_\mathrm{rec} + \lambda_\mathrm{TD} \mathbf{L}_\mathrm{TD}.
	\end{equation}

	\subsubsection{Conflict Detection during Inference}
	During training, we require low-level features to predict their corresponding slots with the projection network $\mathcal{P}$, which provides a method for the model to refine the slots during inference, which we call \textbf{conflict detection (CD)}: Ideally, the prediction of a low-level feature should be close to its corresponding slot. This indicates two facts: (i) When a feature's prediction is far away from all the slots, it represents a wrong perception, such as an undiscovered object, and (ii) When a feature's prediction is close to multiple slots, these slots may represent a single object divided into multiple parts. We resolve such conflicts by adding and merging slots, thereby improving the object-centric representations.
	
	Specifically, we first solve the first issue. After the model extracts low-level features $\mathbf{F}$ and slots $\mathbf{S}$ from the images with the bottom-up pathway, we use the projection network $\mathcal{P}(\cdot)$ to produce a prediction $\mathcal{P}(\mathbf{F})$ and computes the cosine distance between $\mathbf{S}$ and $\mathcal{P}(\mathbf{F})$ and acquire the conflict $\mathbf{C}$. Here we define $\mathbf{C}$ as the distance from $\mathcal{P}(\mathbf{F})$ to their nearest slot:
	
	\begin{equation}
		\mathbf{C} := \mathrm{min}(1 - \mathrm{CosSim}(\mathcal{P}(\mathbf{F}), \mathbf{S})).
	\end{equation}
	We set a threshold $\mathrm{th}$ to determine whether a conflict is large or not. Repetitively, we select $\hat{f}$ from $\mathbf{F}$ that has the largest conflict. If the conflict of $\hat{f}$ exceeds $\mathrm{th}$, we add $\mathcal{P}(\hat{f})$ to the slot set. This process is repeated until all the conflicts are lower than $\mathrm{th}$. After that, we use an agglomerative clustering algorithm to merge slots with less cosine distance than $\mathrm{th}$, thus mitigating the situation where an object is divided into multiple parts. As for the choice of $\mathrm{th}$, we propose a heuristic method that for each trained model, we calculate the average distance between slots and set $\mathrm{th}$ as half of this distance.

	\section{Experiments}
	
	In this section, we first evaluate the object-centric representations with the object discovery task, demonstrating TDGNet's superior object discovery performance across multiple scenes of varying complexity. We then evaluate the performance of the learned object-centric representations in downstream tasks such as unconditional/conditional video prediction and video prediction for visual planning. Specifically, for conditional video prediction and visual planning, we introduce robotics benchmarks~\cite{robonet,vp2} to demonstrate that TDGNet's object-centric representations can be applied in a variety of scenarios.

	\subsection{Unsupervised Object Discovery}
	
	\textbf{Setup.} We first introduce the object discovery task for evaluating object-centric representations. We use three common datasets: CLEVRTex~\cite{https://doi.org/10.48550/arxiv.2111.10265}, MOVi-C~\cite{movi}, and COCO~\cite{coco}. The model's generalization ability is also evaluated using CLEVRTex-OOD and -CAMO, two out-of-distribution test sets. The complexity of the datasets varies. Objects in CLEVRTex have regular shapes, but the challenge is that it adds texture maps to objects and backgrounds, increasing the appearance complexity of objects. MOVi-C takes a further step to use realistic, richly textured objects from the GSO dataset~\cite{downs2022googlescannedobjectshighquality} to create multi-object scenes. COCO contains a large number of natural scene images, and the variability of object appearance has increased significantly. Following previous works~\cite{https://doi.org/10.48550/arxiv.2111.10265}, we use the foreground adjusted rand index (ARI-FG)~\cite{rand1971objective}, mean IoU (mIoU), and mean square error (MSE) to evaluate the models' ability to discover objects and reconstruct the images. For MOVi-C and COCO, we follow~\cite{jiang2023objectcentric,seitzer2022bridging} to include mean best overlapping (mBO) for evaluation.
	
	\noindent \textbf{Result.} Table \ref{tab1} displays quantitative comparison results on CLEVRTex and MOVi-C. In the comparison, we adopt two inference modes of TDGNet: one that only initiates the bottom-up pathway and another that applies the conflict detection (CD) operation to acquire a refined perception. TDGNet with only the bottom-up pathway, outperforms most current models. Conflict detection provides a further performance boost. TDGNet also generalizes well on CLEVRTex-OOD and -CAMO, outperforming current SOTA models by a large margin. 
	
\begin{figure}[t]
		\centering
		\includegraphics[width=1.0\linewidth]{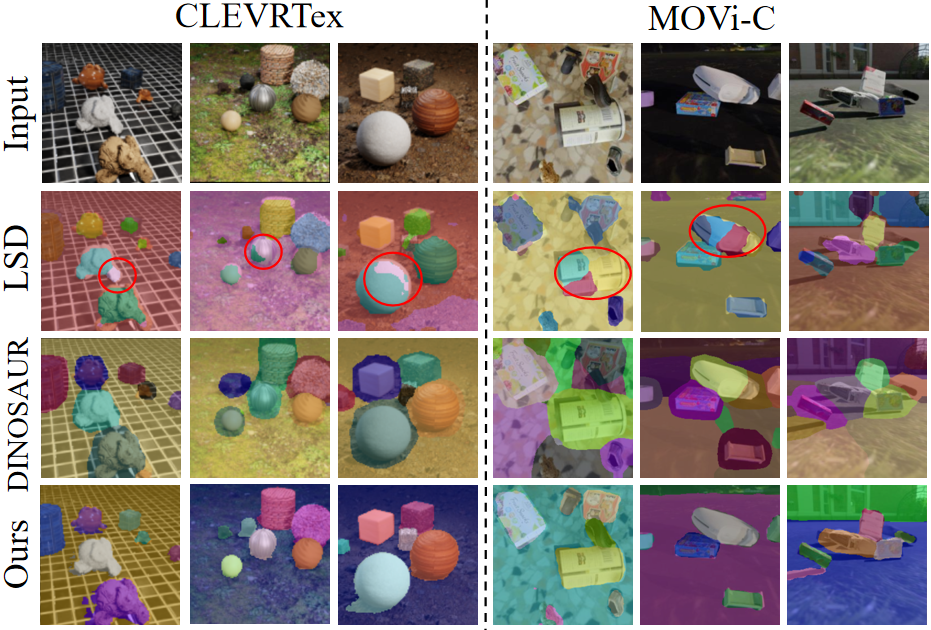}
		\caption{Segmentation results on CLEVRTex and MOVi. TDGNet successfully distinguishes the image's background while segmenting objects with more correct boundaries.}
		\label{fig: segmentation}
	\end{figure}

	On MOVi-C, SOTA models, such as LSD and DINOSAUR, rely on reconstructing features from pre-training backbones to generalize to complex scenarios. However, these models show significant biases in different evaluation metrics. For instance, DINOSAUR offers an ARI-FG far superior to other models but provides lower mIoU and mBO. On the other hand, the performance of LSD is just the opposite, with lower ARI-FG and higher mIoU and mBO. Our TDGNet, by contrast, outperforms previous models in all the metrics, demonstrating a more comprehensive object discovery capability. Figure \ref{fig: segmentation} gives examples to illustrate the improvement. DINOSAUR fails to fit the edges and cannot segment the background holistically on MOVi-C. LSD tends to divide objects into multiple parts as is circled. TDGNet, instead, succeeds in segmenting backgrounds, as well as showing better object understanding where the object masks fit the outlines of objects well and do not divide objects into parts, thus achieving higher scores on all metrics.
	
	We further provide the comparison result on the COCO dataset in Table \ref{tab2}. Following~\cite{seitzer2022bridging}, we discuss the situations when different types of decoders are used. For the MLP-based decoder, we adopt DINOSAUR as the baseline and use DINO\_v2 as the backbone. Compared with DINOSAUR trained only through reconstruction, we have improved by 2.1, 0.7, and 1.2 respectively, according to ARI-FG, mBO, and mIoU. For the transformer-based decoder, we choose SPOT as the baseline and combine it with TDGNet, achieving performance improvements of 0.3, 0.6, and 0.2 in terms of the three metrics.


	\subsection{Exploring Object-Centric Representation in Predicting the World's State}
	
	We believe that object-centric learning, which uses "objects" as the basic unit to represent visual scenes, is worth exploring in world modeling. Here we explore the field of robotics, introducing video prediction and visual planning tasks to evaluate whether the object-centric representations from TDGNet benefit the prediction of the world's states. In video prediction tasks, after training TDGNet, we additionally train an auto-regressive Transformer to predict the future slots, and then decode these slots to acquire future frames. Furthermore, the visual planning task evaluates the video prediction model by using the prediction results in control tasks.
	
	\begin{table}[t]
		\small
		\begin{center}

			\setlength{\tabcolsep}{0.3mm}
			\begin{tabular}{l c c c} 
				\toprule[2pt]
				\makebox[0.26\textwidth][c]{\textbf{MOVi-C}} &
				$\uparrow$PSNR & $\uparrow$SSIM & $\downarrow$LPIPS \\
				\hline
				SlotFormer~\cite{wu2023slotformerunsupervisedvisualdynamics}           
				& 19.5 & 45.6 & 53.4 \\
				OCVP~\cite{villar2023object}                 
				& 19.9 & 50.2 & 45.0 \\
				OCK~\cite{song2024unsupervised}                  
				& \underline{21.0} & \underline{59.3} & \underline{37.0} \\
				\textbf{TDGNet (ours)} & \textbf{22.6} & \textbf{66.3} & \textbf{25.9} \\
				\bottomrule[2pt]
			\end{tabular}
			\caption{Unconditional video prediction on MOVi-C. LPIPS and SSIM scores are scaled by 100 for convenient display.}
			
			\label{tab:video_prediction}
		\end{center}
		
	\end{table}

	\begin{table}[t]
		\small
		\begin{center}

			\setlength{\tabcolsep}{0.7mm}
			\begin{tabular}{l c c c c} 
				\toprule[2pt]
				\makebox[0.23\textwidth][c]{\textbf{RoboNet}} &
				\makebox[0.05\textwidth][c]{$\downarrow$FVD} & 
				\makebox[0.05\textwidth][c]{$\uparrow$PSNR} & 
				\makebox[0.05\textwidth][c]{$\uparrow$SSIM} & 
				\makebox[0.05\textwidth][c]{$\downarrow$LPIPS} \\
				\hline
				
				MaskViT~\cite{gupta2022maskvitmaskedvisualpretraining} 
				& 211.7 & 20.4 & 67.1 & 17.0 \\
				iVideoGPT~\cite{wu2024ivideogptinteractivevideogptsscalable}
				& \underline{197.9} & \underline{23.8} & \textbf{80.8} & \underline{14.7} \\
				\textbf{TDGNet (ours)} 
				& \textbf{187.3} & \textbf{23.9} & \underline{79.7} & \textbf{14.2} \\
				\bottomrule[2pt]
			\end{tabular}
			\caption{Conditional video prediction results on RoboNet. LPIPS and SSIM scores are scaled by 100 for convenient display.}
			
			\label{tab: RoboNet}
		\end{center}
		
	\end{table}

	\subsubsection{Video Prediction}
	
	\noindent \textbf{Setup.} We conduct video prediction on the MOVi-C and RoboNet datasets. MOVi-C, a widely used multi-object dataset, contains videos of the interactions of moving objects. Following previous works~\cite{song2024unsupervised}, the model predicts 8 future frames from 6 context frames with no conditions. We primarily compare TDGNet to existing Object-Centric models in MOVi-C to ensure that the slots provided by TDGNet outperform existing models. RoboNet is a large-scale robot dataset that contains a large number of videos of object manipulation with robotic arms. The models are required to predict 10 future frames given 2 context frames and the robotic arm's action as conditions. We compare TDGNet to existing conditional video prediction models on RoboNet. Following previous work, we use SSIM, PSNR, and LPIPS for evaluation on MOVi-C, and additionally introduce FVD for RoboNet.
	
\begin{figure}[t]
		\begin{center}
			\includegraphics[width=0.955\linewidth]{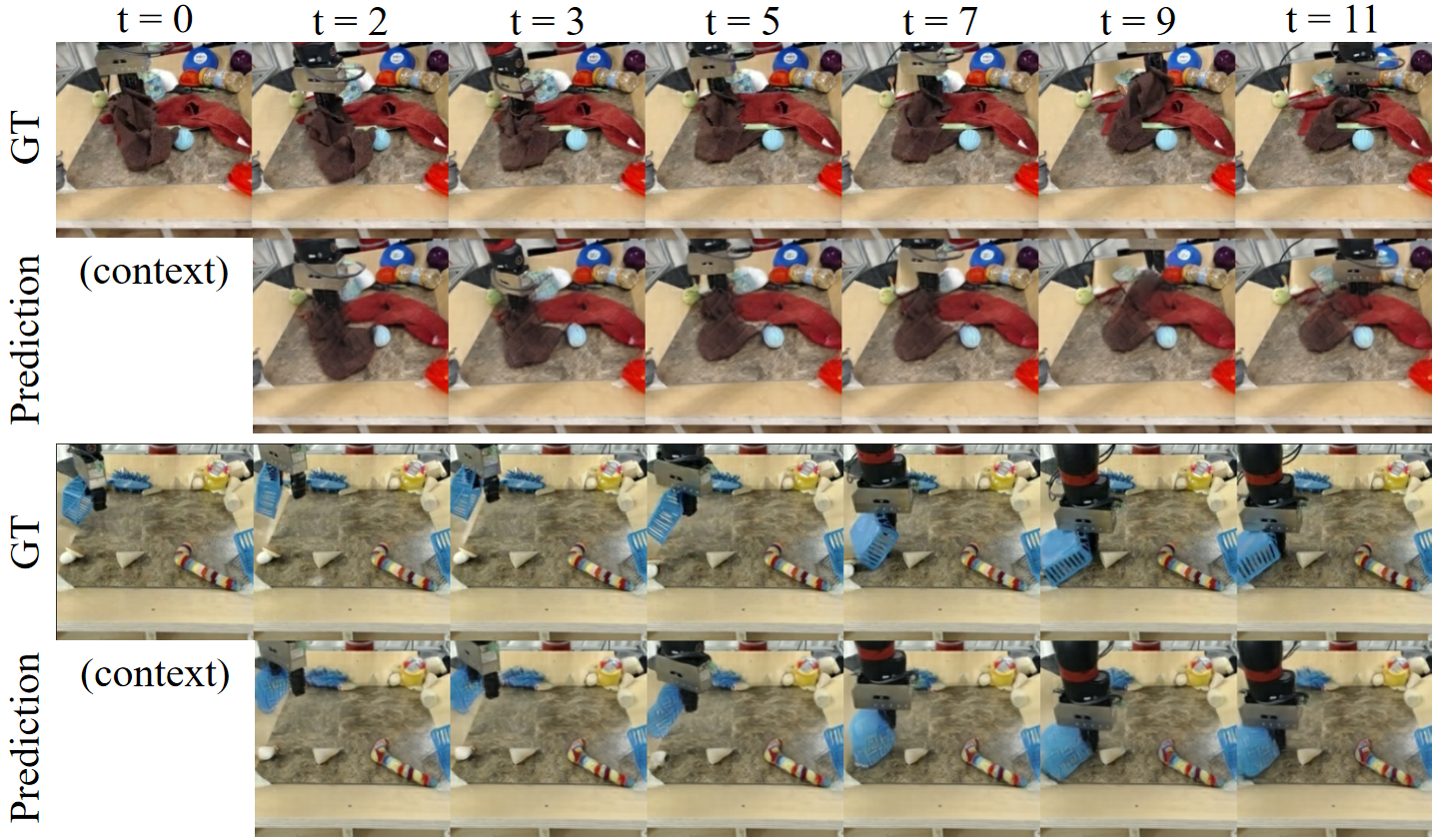}
		\end{center}
		\caption{Video Prediction on RoboNet.} 
		\label{fig: RoboNet}
	\end{figure}

	\noindent \textbf{Result.} We list the comparison results in Table \ref{tab:video_prediction} and \ref{tab: RoboNet}. For MOVi-C, using the slots extracted by TDGNet for prediction significantly improves the performance, outperforming other models by a large margin. For RoboNet, we provide results on 256 resolution. We observe that our model outperforms iVideoGPT, the current SOTA model, in three out of the four evaluation metrics.
	In Figure \ref{fig: RoboNet}, we provide the visualization results on RoboNet, demonstrating that our model accurately simulates the interaction between robots and objects.
	
	\begin{table}[t!]
		\begin{center}
		
			\setlength{\tabcolsep}{1mm}
			\small
			\begin{tabular}{c c c c c c c c} 
				\toprule[2pt]
				\multirow{2}{*}{\textbf{Dataset}}  & 
				\multicolumn{4}{c}{\textbf{Components}} & 
				\multicolumn{3}{c}{\textbf{Metrics}} \\
				\cmidrule(r){2-5}\cmidrule(r){6-8}
				& \makebox[0.03\textwidth][c]{L1}
				& \makebox[0.03\textwidth][c]{LP} 
				& \makebox[0.03\textwidth][c]{TDG}
				& \makebox[0.03\textwidth][c]{CD}
				& \makebox[0.05\textwidth][c]{ARI-FG}
				& \makebox[0.05\textwidth][c]{mBO} 
				& \makebox[0.05\textwidth][c]{mIoU} \\
				\hline
				\multirow{4}{*}{CLEVRTex} 
				& \checkmark & & & & 88.3 & 74.2 & 73.1 \\
				& \checkmark & \checkmark & & & 90.9 & 78.4 & 77.6 \\
				& \checkmark & \checkmark & \checkmark & & 94.2 & 80.4 & 80.3 \\
				& \checkmark & \checkmark & \checkmark & \checkmark & \textbf{94.8} & \textbf{80.8} & \textbf{80.5} \\
				\hline
				\multirow{4}{*}{MOVi-C} 
				& \checkmark & & & & 52.9 & 36.4 & 33.1 \\
				& \checkmark & \checkmark & & & 58.6 & 46.8 & 44.9 \\
				& \checkmark & \checkmark & \checkmark & & 61.2 & 53.5 & 52.9 \\
				& \checkmark & \checkmark & \checkmark & \checkmark & \textbf{68.5} & \textbf{57.1} & \textbf{55.6} \\
				\bottomrule[2pt]
			\end{tabular}
			\caption{Ablation on the components of TDGNet. `L1' and `LP' represent the L1 and LPIPS loss used for reconstruction, while `TDG' and `CD' represent top-down guidance and conflict detection.}
			
			\label{tab: flexibility}
		\end{center}
	\end{table}
	
	\begin{figure*}[t]
		\centering
		\includegraphics[width=1.0\linewidth]{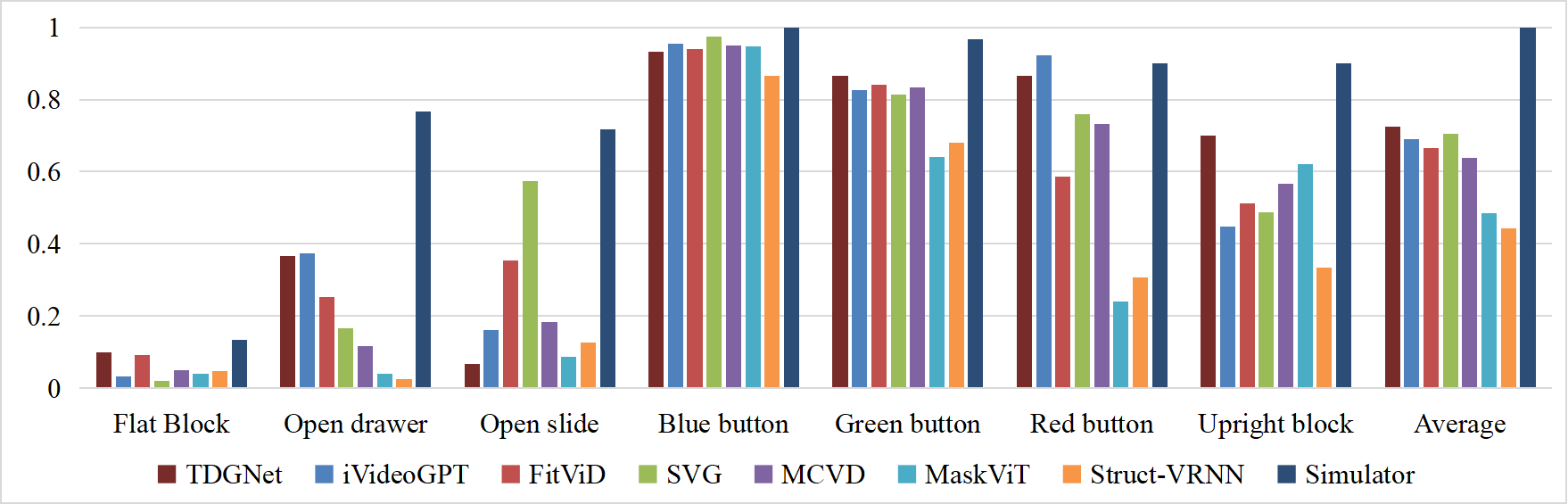}
		\caption{Visual planning on the $\mathrm{VP}^2$ benchmark. On the right, we show the mean scores for each model averaged across
			all tasks, normalized by the performance of the simulator.}
		\label{fig: planning}
	\end{figure*}
	\subsubsection{Visual Planning}
	
	\noindent \textbf{Setup.} As is discussed in previous works~\cite{vp2}, the video prediction performance sometimes cannot reflect whether models correctly predict the world's state. Therefore, we use their proposed $\mathrm{VP}^2$ benchmark~\cite{vp2} to evaluate the performance of TDGNet in the visual planning task. In $\mathrm{VP}^2$, a given agent uses the predicted frames from video prediction models to complete various control tasks. All elements are provided except the video prediction models, thus comparing them under a standardized benchmark. We use the RoboDesk environment in $\mathrm{VP}^2$ for evaluation, which includes 7 kinds of control tasks. The agent's success rate to complete the tasks is used as the evaluation metric.
	
	\noindent \textbf{Result.} In Figure \ref{fig: planning}, we present a comparison between TDGNet and a set of baseline models~\cite{wu2024ivideogptinteractivevideogptsscalable,babaeizadeh2021fitvidoverfittingpixellevelvideo,villegas2019highfidelityvideoprediction,voleti2022mcvdmaskedconditionalvideo,minderer2020unsupervisedlearningobjectstructure,gupta2022maskvitmaskedvisualpretraining}. `Simulator' is the success rate of the agent when it directly use the simulator as the dynamics, representing an upper bound of model performance. TDGNet attains performance that is comparable or superior to existing SOTA models across the majority of tasks. In addition, we compute the average success rate, normalized by the simulator's performance, demonstrating that TDGNet outperforms existing SOTA models.

	\subsection{Ablative Experiments}
	\label{Ablation}
	
	We conduct ablative experiments on the components of TDGNet during training and inference, including the L1 and LPIPS loss used for reconstruction, as well as the top-down guidance (TDG) and conflict detection (CD) performed by the top-down pathway during training and inference. According to the result in Table \ref{tab: flexibility}, the LPIPS loss improves the performance of baseline models trained only with L1 loss, particularly for MOVi-C, with improvements of 5.7, 10.4, and 11.8, respectively, in terms of ARI-FG, mBO, and mIoU. The top-down pathway further provides considerable performance gains. TDG and CD bring about improvements of 3.9, 2.4, and 2.9 on CLEVRTex, as well as 9.9, 10.3, and 10.7 on MOVi-C. The top-down pathway offers greater improvements on more complex multi-object scenes (i.e., MOVi-C), indicating that the top-down pathway in TDGNet tends to help the model overcome more challenging samples.

	\subsection{Analysis}
	\subsubsection{Low-level Feature Optimization with TDG}
\begin{figure}[t]
		\centering
		\subfloat[]{\includegraphics[width=0.678\columnwidth]{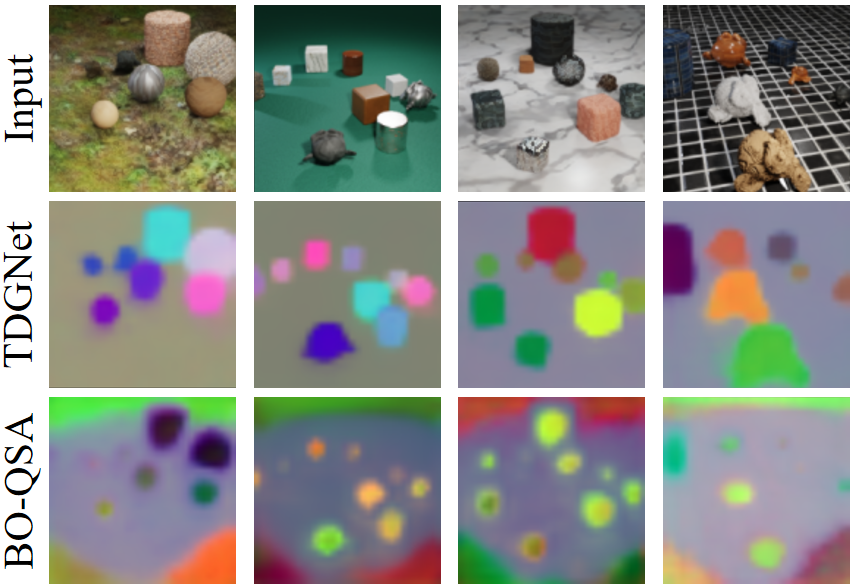}}\hspace{1pt}
		\subfloat[]{\includegraphics[width=0.308\columnwidth]{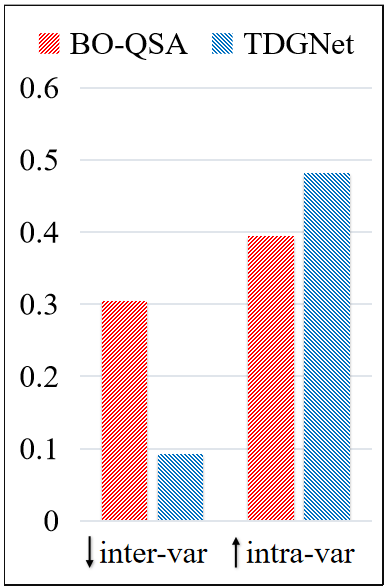}}
		\caption{\textbf{(a)} Feature visualization on CLEVRTex. We use PCA to reduce dimensions to 3. \textbf{(b)} Comparison of object feature variance between TDGNet and BO-QSA. TDGNet has smaller intra- and larger inter-object feature variance.}
		\label{fig: variance}
	\end{figure}

	We visualize the internal features of models in Figure \ref{fig: variance}(a) by taking the low-level features from TDGNet and the baseline model without TDG (i.e., BO-QSA~\cite{jia2023improving}), using PCA~\cite{abdi2010principal} to reduce the feature dimension to 3 for visualization.
	Feature maps from BO-QSA are fuzzy. Some of the small objects are almost unseen in the feature map. 
	By contrast, the features extracted by TDGNet are more conducive to identifying individual objects. The features of the same objects are highly similar, while definite boundaries separate adjacent objects. Further in Figure \ref{fig: variance}(b), we calculate the intra- and inter-object feature variances, which confirms that TDGNet achieves a higher inter-object feature variance and a much lower intra-object feature variance, making objects more distinguishable in the low-level features.
	
	\subsubsection{Iterative Refinement with CD}
\begin{figure}[t]
		\begin{center}
			\includegraphics[width=0.955\linewidth]{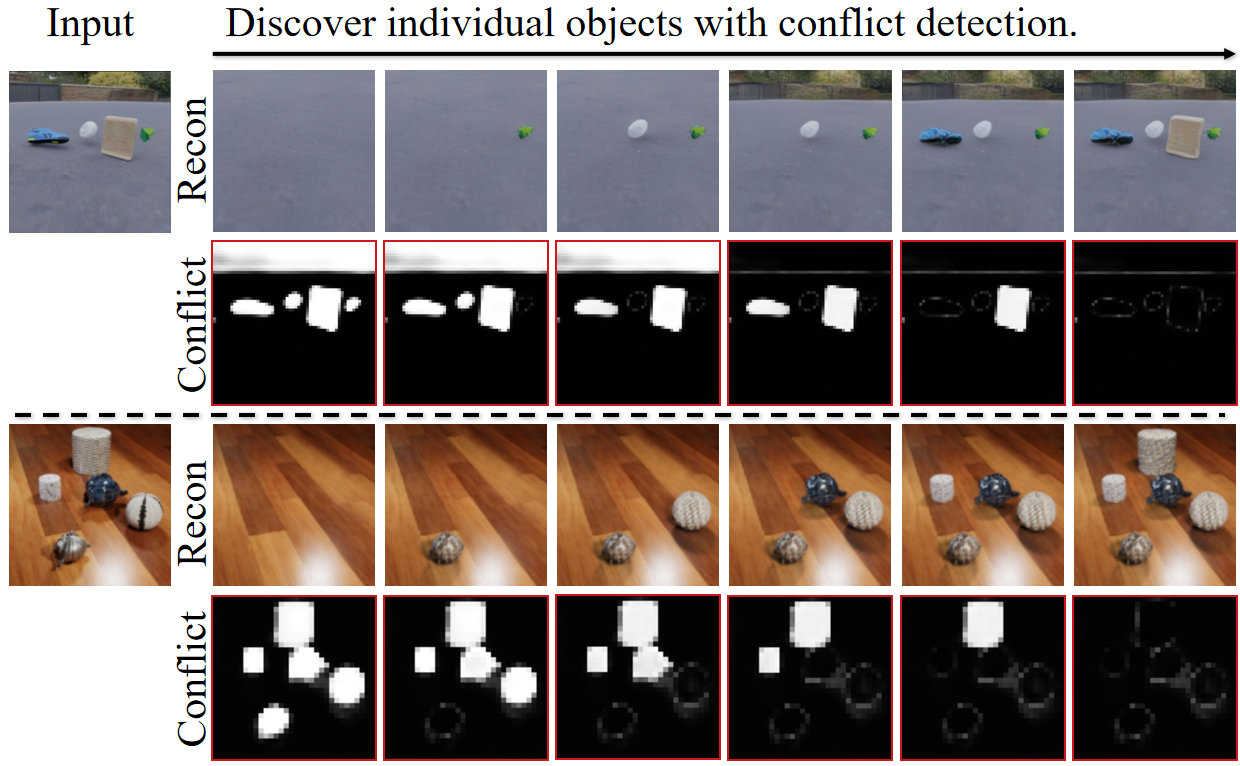}
		\end{center}
		\vspace{-0.5 em}
		\caption{Refinement process of Conflict Detection. We remove all slots except for the background and gradually incorporate objects into the network through CD, and eventually discover all the objects.} 
		\label{fig: Infer-RHG}
		
	\end{figure}

	Compared with extracting objects directly through attention competition, the process of detecting and resolving conflicts provides a more explainable way of object discovery. We visualize the process of CD in Figure \ref{fig: Infer-RHG}. The model repeatedly computes conflicts between low-level features and slots and includes features with large conflicts into the slot set. To clearly illustrate this process, we manually corrupt the bottom-up pathway, assuming that no object is detected and only the background is given at first. In this extreme case, CD repeatedly calculates the conflicts and includes objects. After multiple iterations, all conflicts are solved, representing that all objects are included.

	\section{Conclusion}
	
	Observing that OCL models provide suboptimal object-centric representations such as missing objects or splitting objects into parts, we analyze existing models and propose that OCL models trained solely with reconstruction loss cannot learn distinguishable low-level features. To address this issue, we refer to human vision and propose TDGNet, which introduces a top-down pathway that guides low-level features with the high-level representations (i.e., slots). We verify that the top-down pathway makes objects more distinguishable in the low-level features. Our experiment results show that TDGNet outperforms existing SOTA models in the object discovery task while also demonstrating its potential for downstream tasks such as prediction and planning in the robotics field.
	
	\appendix
	
	\section*{Acknowledgments}
	
	This work was supported by Chinese National Natural Science Foundation Projects U23B2054, 62276254, 62206280, 62376265, the Beijing Science and Technology Plan Project Z231100005923033, Beijing Natural Science Foundation L221013, L242092, the Strategic Priority Research Program of Chinese Academy of Sciences under Grant XDA0480103, the Youth Innovation Promotion Association CAS Y2021131 and InnoHK program.

	\bibliographystyle{named}
	\bibliography{ijcai25}

\begin{thebibliography}{}

\bibitem[\protect\citeauthoryear{Abdi and Williams}{2010}]{abdi2010principal}
Herv{\'e} Abdi and Lynne~J Williams.
\newblock Principal component analysis.
\newblock {\em Wiley interdisciplinary reviews: computational statistics},
  2(4):433--459, 2010.

\bibitem[\protect\citeauthoryear{Ahissar and
  Hochstein}{2004}]{ahissar2004reverse}
Merav Ahissar and Shaul Hochstein.
\newblock The reverse hierarchy theory of visual perceptual learning.
\newblock {\em Trends in cognitive sciences}, 8(10):457--464, 2004.

\bibitem[\protect\citeauthoryear{Anderson \bgroup \em et al.\egroup
  }{2018}]{anderson2018bottom}
Peter Anderson, Xiaodong He, Chris Buehler, Damien Teney, Mark Johnson, Stephen
  Gould, and Lei Zhang.
\newblock Bottom-up and top-down attention for image captioning and visual
  question answering.
\newblock In {\em Proceedings of the IEEE conference on computer vision and
  pattern recognition}, pages 6077--6086, 2018.

\bibitem[\protect\citeauthoryear{Babaeizadeh \bgroup \em et al.\egroup
  }{2021}]{babaeizadeh2021fitvidoverfittingpixellevelvideo}
Mohammad Babaeizadeh, Mohammad~Taghi Saffar, Suraj Nair, Sergey Levine, Chelsea
  Finn, and Dumitru Erhan.
\newblock Fitvid: Overfitting in pixel-level video prediction, 2021.

\bibitem[\protect\citeauthoryear{Beck and Kastner}{2009}]{beck2009top}
Diane~M Beck and Sabine Kastner.
\newblock Top-down and bottom-up mechanisms in biasing competition in the human
  brain.
\newblock {\em Vision research}, 49(10):1154--1165, 2009.

\bibitem[\protect\citeauthoryear{Biza \bgroup \em et al.\egroup
  }{2023}]{biza2023invariant}
Ondrej Biza, Sjoerd van Steenkiste, Mehdi S.~M. Sajjadi, Gamaleldin~F. Elsayed,
  Aravindh Mahendran, and Thomas Kipf.
\newblock Invariant slot attention: Object discovery with slot-centric
  reference frames, 2023.

\bibitem[\protect\citeauthoryear{Burgess \bgroup \em et al.\egroup
  }{2019}]{burgess2019monet}
Christopher~P Burgess, Loic Matthey, Nicholas Watters, Rishabh Kabra, Irina
  Higgins, Matt Botvinick, and Alexander Lerchner.
\newblock Monet: Unsupervised scene decomposition and representation.
\newblock {\em arXiv preprint arXiv:1901.11390}, 2019.

\bibitem[\protect\citeauthoryear{Caesar \bgroup \em et al.\egroup
  }{2018}]{coco}
Holger Caesar, Jasper Uijlings, and Vittorio Ferrari.
\newblock Coco-stuff: Thing and stuff classes in context.
\newblock In {\em Computer vision and pattern recognition (CVPR), 2018 IEEE
  conference on}. IEEE, 2018.

\bibitem[\protect\citeauthoryear{Caron \bgroup \em et al.\egroup
  }{2021}]{caron2021emerging}
Mathilde Caron, Hugo Touvron, Ishan Misra, Hervé Jégou, Julien Mairal, Piotr
  Bojanowski, and Armand Joulin.
\newblock Emerging properties in self-supervised vision transformers, 2021.

\bibitem[\protect\citeauthoryear{Chang \bgroup \em et al.\egroup
  }{2022}]{https://doi.org/10.48550/arxiv.2207.00787}
Michael Chang, Thomas~L. Griffiths, and Sergey Levine.
\newblock Object representations as fixed points: Training iterative refinement
  algorithms with implicit differentiation, 2022.

\bibitem[\protect\citeauthoryear{Dasari \bgroup \em et al.\egroup
  }{2020}]{robonet}
Sudeep Dasari, Frederik Ebert, Stephen Tian, Suraj Nair, Bernadette Bucher,
  Karl Schmeckpeper, Siddharth Singh, Sergey Levine, and Chelsea Finn.
\newblock Robonet: Large-scale multi-robot learning, 2020.

\bibitem[\protect\citeauthoryear{Downs \bgroup \em et al.\egroup
  }{2022}]{downs2022googlescannedobjectshighquality}
Laura Downs, Anthony Francis, Nate Koenig, Brandon Kinman, Ryan Hickman, Krista
  Reymann, Thomas~B. McHugh, and Vincent Vanhoucke.
\newblock Google scanned objects: A high-quality dataset of 3d scanned
  household items, 2022.

\bibitem[\protect\citeauthoryear{Engelcke \bgroup \em et al.\egroup
  }{2019}]{genesis}
Martin Engelcke, Adam~R. Kosiorek, Oiwi~Parker Jones, and Ingmar Posner.
\newblock Genesis: Generative scene inference and sampling with object-centric
  latent representations, 2019.

\bibitem[\protect\citeauthoryear{Greff \bgroup \em et al.\egroup
  }{2019}]{greff2019multi}
Klaus Greff, Rapha{\"e}l~Lopez Kaufman, Rishabh Kabra, Nick Watters,
  Christopher Burgess, Daniel Zoran, Loic Matthey, Matthew Botvinick, and
  Alexander Lerchner.
\newblock Multi-object representation learning with iterative variational
  inference.
\newblock In {\em International conference on machine learning}, pages
  2424--2433. PMLR, 2019.

\bibitem[\protect\citeauthoryear{Greff \bgroup \em et al.\egroup }{2022}]{movi}
Klaus Greff, Francois Belletti, Lucas Beyer, Carl Doersch, Yilun Du, Daniel
  Duckworth, David~J. Fleet, Dan Gnanapragasam, Florian Golemo, Charles
  Herrmann, Thomas Kipf, Abhijit Kundu, Dmitry Lagun, Issam Laradji, Hsueh-Ti,
  Liu, Henning Meyer, Yishu Miao, Derek Nowrouzezahrai, Cengiz Oztireli,
  Etienne Pot, Noha Radwan, Daniel Rebain, Sara Sabour, Mehdi S.~M. Sajjadi,
  Matan Sela, Vincent Sitzmann, Austin Stone, Deqing Sun, Suhani Vora, Ziyu
  Wang, Tianhao Wu, Kwang~Moo Yi, Fangcheng Zhong, and Andrea Tagliasacchi.
\newblock Kubric: A scalable dataset generator, 2022.

\bibitem[\protect\citeauthoryear{Gupta \bgroup \em et al.\egroup
  }{2022}]{gupta2022maskvitmaskedvisualpretraining}
Agrim Gupta, Stephen Tian, Yunzhi Zhang, Jiajun Wu, Roberto Martín-Martín,
  and Li~Fei-Fei.
\newblock Maskvit: Masked visual pre-training for video prediction, 2022.

\bibitem[\protect\citeauthoryear{Hochstein and
  Ahissar}{2002}]{hochstein2002view}
Shaul Hochstein and Merav Ahissar.
\newblock View from the top: Hierarchies and reverse hierarchies in the visual
  system.
\newblock {\em Neuron}, 36(5):791--804, 2002.

\bibitem[\protect\citeauthoryear{Jia \bgroup \em et al.\egroup
  }{2023}]{jia2023improving}
Baoxiong Jia, Yu~Liu, and Siyuan Huang.
\newblock Improving object-centric learning with query optimization.
\newblock In {\em The Eleventh International Conference on Learning
  Representations}, 2023.

\bibitem[\protect\citeauthoryear{Jiang \bgroup \em et al.\egroup
  }{2023}]{jiang2023objectcentric}
Jindong Jiang, Fei Deng, Gautam Singh, and Sungjin Ahn.
\newblock Object-centric slot diffusion, 2023.

\bibitem[\protect\citeauthoryear{Kahneman \bgroup \em et al.\egroup
  }{1992}]{kahneman1992reviewing}
Daniel Kahneman, Anne Treisman, and Brian~J Gibbs.
\newblock The reviewing of object files: Object-specific integration of
  information.
\newblock {\em Cognitive psychology}, 24(2):175--219, 1992.

\bibitem[\protect\citeauthoryear{Kakogeorgiou \bgroup \em et al.\egroup
  }{2024}]{Kakogeorgiou_2024_CVPR}
Ioannis Kakogeorgiou, Spyros Gidaris, Konstantinos Karantzalos, and Nikos
  Komodakis.
\newblock Spot: Self-training with patch-order permutation for object-centric
  learning with autoregressive transformers.
\newblock In {\em Proceedings of the IEEE/CVF Conference on Computer Vision and
  Pattern Recognition (CVPR)}, pages 22776--22786, June 2024.

\bibitem[\protect\citeauthoryear{Karazija \bgroup \em et al.\egroup
  }{2021}]{https://doi.org/10.48550/arxiv.2111.10265}
Laurynas Karazija, Iro Laina, and Christian Rupprecht.
\newblock Clevrtex: A texture-rich benchmark for unsupervised multi-object
  segmentation, 2021.

\bibitem[\protect\citeauthoryear{Karimi-Rouzbahani \bgroup \em et al.\egroup
  }{2017}]{karimi2017invariant}
Hamid Karimi-Rouzbahani, Nasour Bagheri, and Reza Ebrahimpour.
\newblock Invariant object recognition is a personalized selection of invariant
  features in humans, not simply explained by hierarchical feed-forward vision
  models.
\newblock {\em Scientific reports}, 7(1):14402, 2017.

\bibitem[\protect\citeauthoryear{Liang and Hu}{2015}]{liang2015recurrent}
Ming Liang and Xiaolin Hu.
\newblock Recurrent convolutional neural network for object recognition.
\newblock In {\em Proceedings of the IEEE conference on computer vision and
  pattern recognition}, pages 3367--3375, 2015.

\bibitem[\protect\citeauthoryear{Liu \bgroup \em et al.\egroup
  }{2024}]{liu2024primitivenet}
Chang Liu, Xudong Jiang, and Henghui Ding.
\newblock Primitivenet: decomposing the global constraints for referring
  segmentation.
\newblock {\em Visual Intelligence}, 2(1):16, 2024.

\bibitem[\protect\citeauthoryear{Locatello \bgroup \em et al.\egroup
  }{2020}]{https://doi.org/10.48550/arxiv.2006.15055}
Francesco Locatello, Dirk Weissenborn, Thomas Unterthiner, Aravindh Mahendran,
  Georg Heigold, Jakob Uszkoreit, Alexey Dosovitskiy, and Thomas Kipf.
\newblock Object-centric learning with slot attention, 2020.

\bibitem[\protect\citeauthoryear{Minderer \bgroup \em et al.\egroup
  }{2020}]{minderer2020unsupervisedlearningobjectstructure}
Matthias Minderer, Chen Sun, Ruben Villegas, Forrester Cole, Kevin Murphy, and
  Honglak Lee.
\newblock Unsupervised learning of object structure and dynamics from videos,
  2020.

\bibitem[\protect\citeauthoryear{Ramanishka \bgroup \em et al.\egroup
  }{2017}]{ramanishka2017top}
Vasili Ramanishka, Abir Das, Jianming Zhang, and Kate Saenko.
\newblock Top-down visual saliency guided by captions.
\newblock In {\em Proceedings of the IEEE conference on computer vision and
  pattern recognition}, pages 7206--7215, 2017.

\bibitem[\protect\citeauthoryear{Rand}{1971}]{rand1971objective}
William~M Rand.
\newblock Objective criteria for the evaluation of clustering methods.
\newblock {\em Journal of the American Statistical association},
  66(336):846--850, 1971.

\bibitem[\protect\citeauthoryear{Seitzer \bgroup \em et al.\egroup
  }{2022}]{seitzer2022bridging}
Maximilian Seitzer, Max Horn, Andrii Zadaianchuk, Dominik Zietlow, Tianjun
  Xiao, Carl-Johann Simon-Gabriel, Tong He, Zheng Zhang, Bernhard
  Sch{\"o}lkopf, Thomas Brox, et~al.
\newblock Bridging the gap to real-world object-centric learning.
\newblock {\em arXiv preprint arXiv:2209.14860}, 2022.

\bibitem[\protect\citeauthoryear{Singh \bgroup \em et al.\egroup
  }{2021}]{slate}
Gautam Singh, Fei Deng, and Sungjin Ahn.
\newblock Illiterate dall-e learns to compose.
\newblock {\em arXiv preprint arXiv:2110.11405}, 2021.

\bibitem[\protect\citeauthoryear{Song \bgroup \em et al.\egroup
  }{2024}]{song2024unsupervised}
Yeon-Ji Song, Suhyung Choi, Jaein Kim, Jin-Hwa Kim, and Byoung-Tak Zhang.
\newblock Unsupervised dynamics prediction with object-centric kinematics.
\newblock {\em arXiv preprint arXiv:2404.18423}, 2024.

\bibitem[\protect\citeauthoryear{Tian \bgroup \em et al.\egroup }{2023}]{vp2}
Stephen Tian, Chelsea Finn, and Jiajun Wu.
\newblock A control-centric benchmark for video prediction, 2023.

\bibitem[\protect\citeauthoryear{Villar-Corrales \bgroup \em et al.\egroup
  }{2023}]{villar2023object}
Angel Villar-Corrales, Ismail Wahdan, and Sven Behnke.
\newblock Object-centric video prediction via decoupling of object dynamics and
  interactions.
\newblock In {\em 2023 IEEE International Conference on Image Processing
  (ICIP)}, pages 570--574. IEEE, 2023.

\bibitem[\protect\citeauthoryear{Villegas \bgroup \em et al.\egroup
  }{2019}]{villegas2019highfidelityvideoprediction}
Ruben Villegas, Arkanath Pathak, Harini Kannan, Dumitru Erhan, Quoc~V. Le, and
  Honglak Lee.
\newblock High fidelity video prediction with large stochastic recurrent neural
  networks, 2019.

\bibitem[\protect\citeauthoryear{Voleti \bgroup \em et al.\egroup
  }{2022}]{voleti2022mcvdmaskedconditionalvideo}
Vikram Voleti, Alexia Jolicoeur-Martineau, and Christopher Pal.
\newblock Mcvd: Masked conditional video diffusion for prediction, generation,
  and interpolation, 2022.

\bibitem[\protect\citeauthoryear{Wen \bgroup \em et al.\egroup
  }{2018}]{wen2018deep}
Haiguang Wen, Kuan Han, Junxing Shi, Yizhen Zhang, Eugenio Culurciello, and
  Zhongming Liu.
\newblock Deep predictive coding network for object recognition.
\newblock In {\em International conference on machine learning}, pages
  5266--5275. PMLR, 2018.

\bibitem[\protect\citeauthoryear{Wolfe}{2021}]{wolfe2021guided2}
Jeremy~M Wolfe.
\newblock Is guided search 6.0 compatible with reverse hierarchy theory.
\newblock {\em Journal of Vision}, 21(9):36--36, 2021.

\bibitem[\protect\citeauthoryear{Wu \bgroup \em et al.\egroup
  }{2023}]{wu2023slotformerunsupervisedvisualdynamics}
Ziyi Wu, Nikita Dvornik, Klaus Greff, Thomas Kipf, and Animesh Garg.
\newblock Slotformer: Unsupervised visual dynamics simulation with
  object-centric models, 2023.

\bibitem[\protect\citeauthoryear{Wu \bgroup \em et al.\egroup
  }{2024}]{wu2024ivideogptinteractivevideogptsscalable}
Jialong Wu, Shaofeng Yin, Ningya Feng, Xu~He, Dong Li, Jianye Hao, and
  Mingsheng Long.
\newblock ivideogpt: Interactive videogpts are scalable world models, 2024.

\bibitem[\protect\citeauthoryear{Yin \bgroup \em et al.\egroup
  }{2022}]{yin2022transfgu}
Zhaoyuan Yin, Pichao Wang, Fan Wang, Xianzhe Xu, Hanling Zhang, Hao Li, and
  Rong Jin.
\newblock Transfgu: a top-down approach to fine-grained unsupervised semantic
  segmentation.
\newblock In {\em Computer Vision--ECCV 2022: 17th European Conference, Tel
  Aviv, Israel, October 23--27, 2022, Proceedings, Part XXIX}, pages 73--89.
  Springer, 2022.

\bibitem[\protect\citeauthoryear{Zhang \bgroup \em et al.\egroup
  }{2018}]{zhang2018unreasonable}
Richard Zhang, Phillip Isola, Alexei~A. Efros, Eli Shechtman, and Oliver Wang.
\newblock The unreasonable effectiveness of deep features as a perceptual
  metric, 2018.

\end{thebibliography}
	
\end{document}